\documentclass[pmlr]{jmlr}

\usepackage{graphicx}
\usepackage{amsmath}
\usepackage{booktabs}
\usepackage{multirow}
\usepackage{array}
\usepackage{xurl} 
\usepackage[compact]{titlesec}

\titlespacing{\section}{0pt}{6pt}{3pt}

\jmlrvolume{--}
\jmlryear{2026}
\jmlrsubmitted{Preprint}
\jmlrworkshop{Preprint}

\title[When is 3D Worth It?]
{When is 3D Worth It? A Resource--Performance Frontier for CNNs and Transformers in Lung CT}

\author{
\Name{Md Enamul Hoq}$^{1}$ \\
\Name{Sharafat Hossain}$^{2}$ \\
\Name{Imraul Emmaka}$^{2}$ \\
\Name{Linda Larson-Prior}$^{3}$ \\
\Name{Lawrence Tarbox}$^{1}$ \\
\Name{Jonathan Bona}$^{1}$ \\
\Name{Donald Johann Jr.}$^{1}$ \\
\Name{Fred Prior}$^{1}$ \\
\\
$^{1}$Department of Biomedical Informatics, University of Arkansas for Medical Sciences \\
$^{2}$Department of Information Science, University of Arkansas at Little Rock \\
$^{3}$Department of Neuroscience, University of Arkansas for Medical Sciences
}

\begin{document}

\maketitle

\begin{abstract}
Three-dimensional models are widely assumed preferable for volumetric medical imaging, yet their practical value depends on whether performance gains justify added computational cost and complexity. Rather than proposing a new architecture, we study how input dimensionality (2D, 2.5D, 3D) affects model behavior across convolutional neural networks (CNNs) and Vision Transformers (ViTs) under a fixed training protocol.
Using a leakage-free NLST cohort ($n=1{,}977$) with supporting LIDC-IDRI data, we find that the 2.5D CNN offers the most favorable discrimination--stability trade-off in our comparison (ROC-AUC 0.682, 95\% CI [0.546, 0.799]) with a stable operating point. In contrast, 3D CNNs show threshold instability, and transformers exhibit degenerate predictions (e.g., all-positive).
Confidence intervals are wide and overlapping, so we present these results as a controlled resource--performance frontier and a failure-mode taxonomy rather than as definitive superiority claims. For class-imbalanced lung cancer screening classification, 2D and 2.5D inputs provide a more reliable trade-off between performance, stability, and computational efficiency than full 3D representations.
\end{abstract}


\section{Introduction}
Deep learning for lung CT often presumes 3D superiority due to volumetric context \citep{ardila2019end,mikhael2023sybil}. Yet volumetric models incur higher memory, longer training, and optimization difficulty \citep{wu2020mdndnet,nasrullah2019automated}. Hybrid 2.5D strategies balance context and efficiency, while transformers add sensitivity to scale and data regime \citep{dosovitskiy2021an,wang2022uctransnet}. Recent lung CT foundation models underscore that strong performance can emerge from carefully designed 2D or hybrid pipelines \citep{hoq2025harnessing,hoq2026virtualeyes,agrawal2025pillar0,veenboer2025tapct}. Despite these trends, the question remains: when is 3D actually worth its cost and added input complexity? We address this via a controlled comparison of 2D, 2.5D, and 3D inputs across CNNs and transformers, evaluating discrimination and failure modes in class-imbalanced lung CT cohorts.

Our contribution is deliberately diagnostic rather than architectural. Holding the training protocol fixed, we (i) map a resource--performance frontier across input dimensionality for both CNN and ViT families, and (ii) characterize the operating-point failures---threshold instability and degenerate all-positive/all-negative predictions---that ROC-AUC alone conceals. We do not claim a new model or state-of-the-art accuracy; we treat the single-cohort scope, modest absolute performance, and wide confidence intervals as explicit limitations (Section~\ref{sec:limit}), and intend this as a controlled, reproducible starting point for larger studies.

\section{Methods}
\textbf{Cohorts and outcome.} We used a leakage-free NLST cohort ($n=1{,}977$) with fixed, patient-level splits (1,426 train, 254 val, 297 test) and a lung-focused QC pipeline (Virtual-Eyes) that standardizes series selection and lung localization, retaining one canonical CT series per patient \citep{hoq2026virtualeyes}. The prediction target is a binary patient-level cancer label derived from the NLST screening and diagnosis records. The task is strongly imbalanced: the held-out test set contains 20 cancer-positive and 277 non-cancer patients ($\approx$6.7\% prevalence), and the splits preserve patient-level prevalence. LIDC-IDRI provided a supporting cohort with XML-derived weak labels, used only for external corroboration of the failure-mode pattern and not for training the NLST models.

\textbf{Inputs and models.} From each canonical volume we construct three inputs that differ only in dimensionality: 2D (a single central axial slice through the lung region), 2.5D (three orthogonal slices stacked as channels), and 3D (a lung-centered sub-volume). Two matched backbone families---a residual CNN and a Vision Transformer (ViT)---are adapted to each input and trained under an identical protocol (20 epochs; weighted binary cross-entropy to address imbalance). Optimizer and learning-rate settings are reported in Appendix~A.1.

\textbf{Evaluation.} We report ROC-AUC, PR-AUC, and sensitivity/specificity at both the default (0.5) and validation-selected thresholds, with bootstrap 95\% confidence intervals for AUC. GPU memory and inference time index computational cost.

\begin{figure}[t]
\centering
\includegraphics[width=\linewidth]{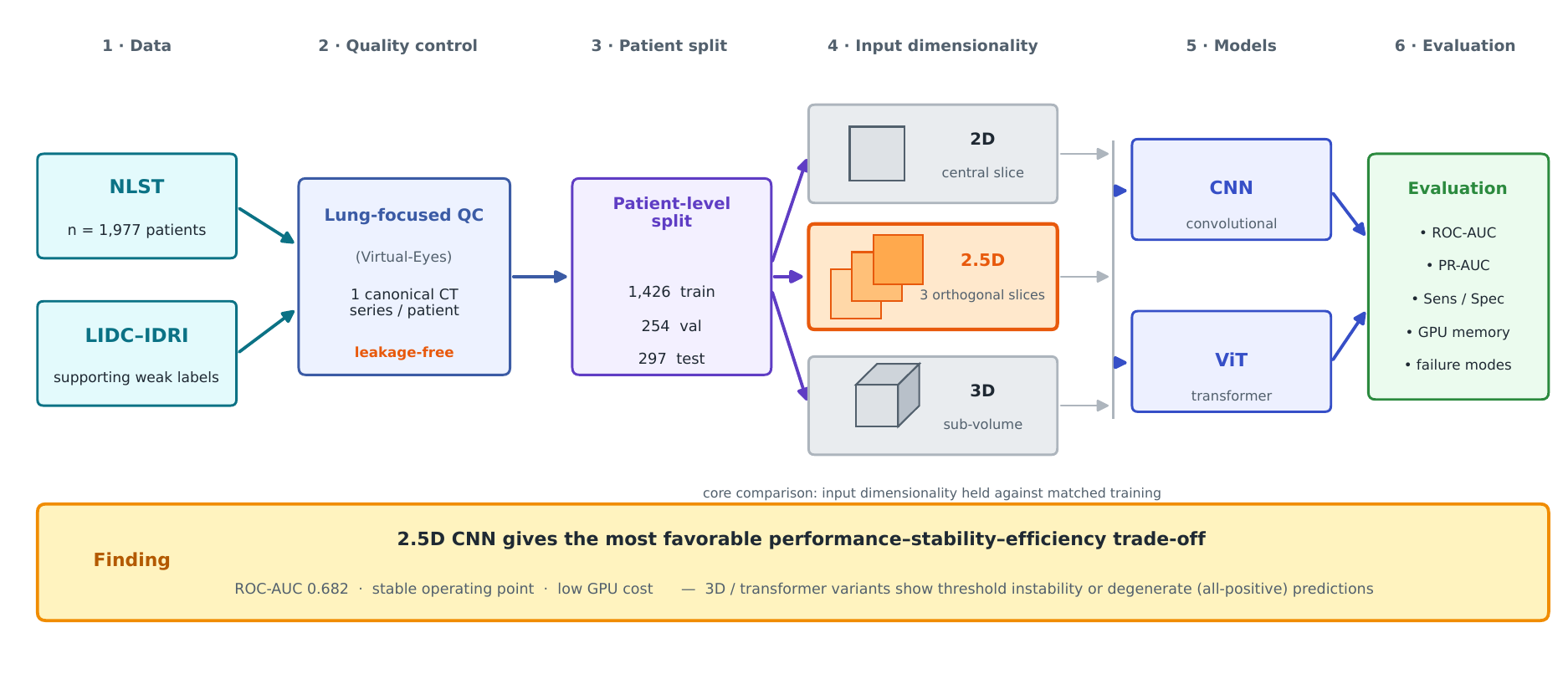}
\caption{Study design and resource--performance frontier. A lung-focused QC pipeline yields one canonical volume per patient; a leakage-free patient-level split feeds three inputs that differ only in dimensionality (2D, 2.5D, 3D) into matched CNN and ViT backbones, which are evaluated for discrimination, operating behavior, and computational cost. The 2.5D CNN gives the most favorable trade-off.}
\label{fig:pipeline}
\end{figure}

\section{Results}
NLST results (Table~\ref{tab:nlst}) show the 2.5D CNN achieved the highest ROC-AUC (0.682) and PR-AUC (0.158) in our comparison. The 2D CNN was overly conservative (sens 0.10, spec 0.949); the 3D CNN exhibited threshold instability (sens 0.05--0.50 across thresholds). Transformers required roughly 3$\times$ GPU memory and frequently collapsed: the 3D ViT produced zero sensitivity despite ROC-AUC 0.589; the 2.5D ViT, while moderately discriminative, showed extreme threshold sensitivity. LIDC-IDRI reproduced the pattern: higher-capacity models predicted all cases as positive (specificity 0) despite moderate AUC. We stress that the bootstrap confidence intervals are wide and overlap substantially across models (Table~\ref{tab:nlst_expanded}); the ranking differences are not statistically separable, so we emphasize operating behavior and computational cost alongside discrimination rather than point-estimate ordering alone.

\begin{table}[t]
\centering
\scriptsize
\resizebox{\linewidth}{!}{%
\begin{tabular}{lccccc}
\toprule
Model & ROC-AUC & PR-AUC & Sens (Def/Val) & Spec (Def/Val) & GPU (MB) \\
\midrule
2D CNN   & 0.581 & 0.088 & 0.10 / 0.10 & 0.949 / 0.931 & 1620 \\
2.5D CNN & \textbf{0.682} & \textbf{0.158} & 0.20 / 0.75 & 0.949 / 0.469 & 1646 \\
3D CNN   & 0.622 & 0.107 & 0.05 / 0.50 & 0.975 / 0.671 & 1777 \\
2D ViT   & 0.598 & 0.088 & 0.10 / 0.10 & 0.910 / 0.881 & 4959 \\
2.5D ViT & 0.631 & 0.127 & 0.10 / 0.60 & 0.986 / 0.505 & 4959 \\
3D ViT   & 0.589 & 0.081 & 0.00 / 0.00 & 1.00 / 0.964 & 352 \\
\bottomrule
\end{tabular}%
}
\caption{NLST results with 95\% CI for the 2.5D CNN: [0.546,0.799]. Sensitivity/specificity at default (0.5) and validation-optimized thresholds. Confidence intervals for all models overlap (Appendix Table~\ref{tab:nlst_expanded}).}
\label{tab:nlst}
\end{table}

\begin{figure}[t]
\centering
\includegraphics[width=0.48\linewidth]{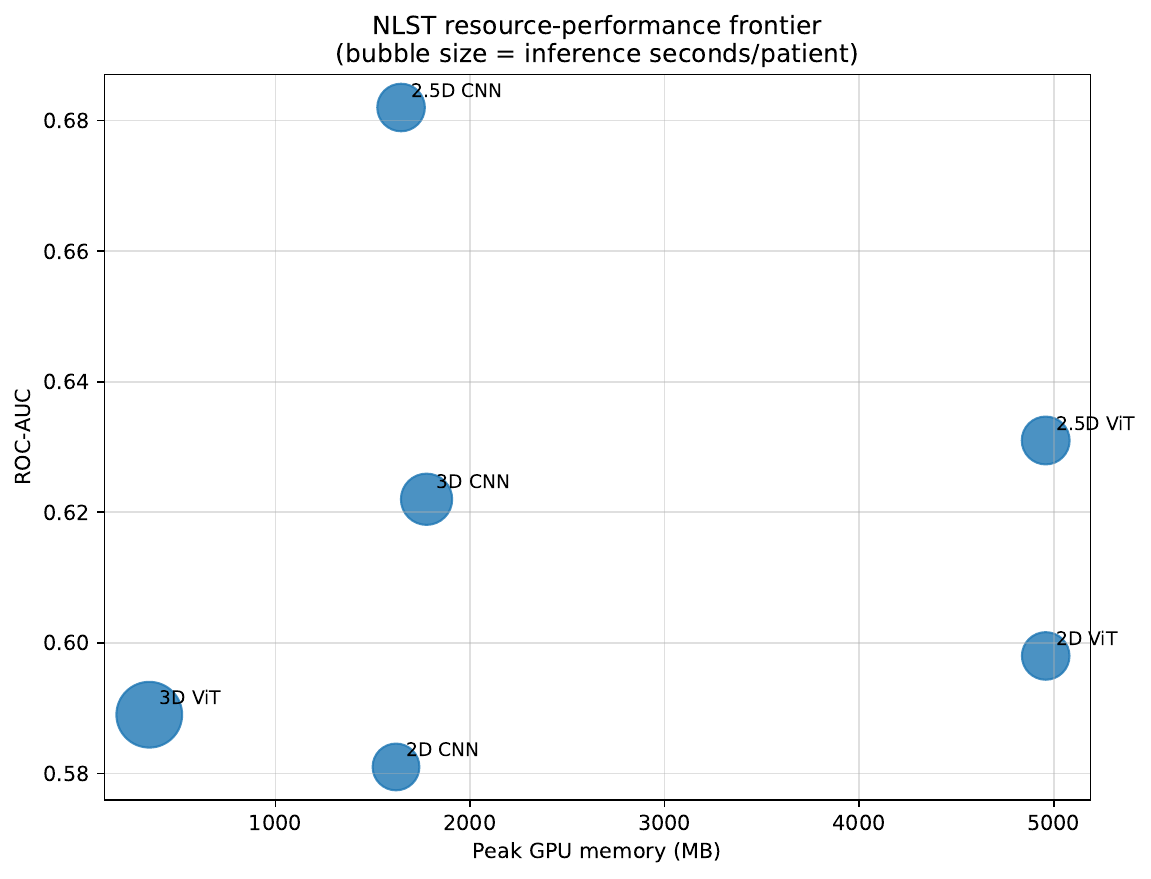}
\includegraphics[width=0.48\linewidth]{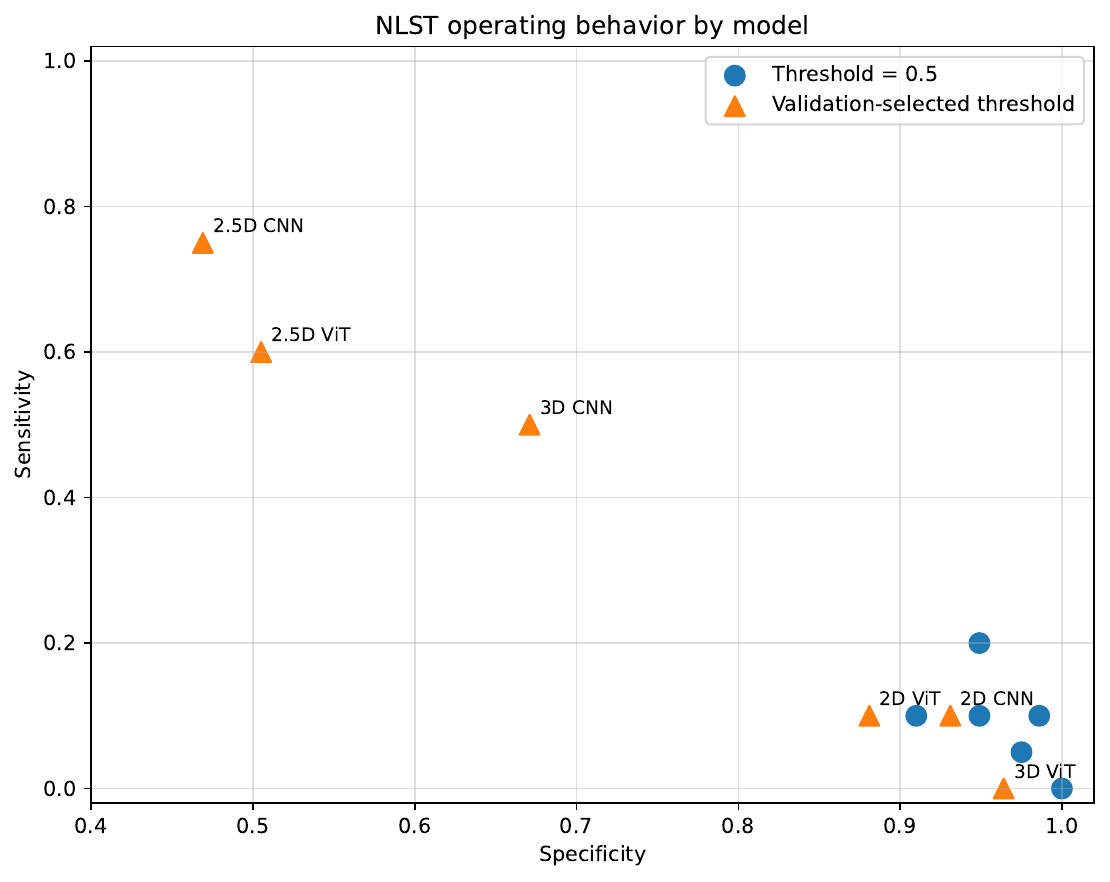}
\caption{Left: Resource--performance frontier (bubble size $\propto$ inference time). Right: Operating behavior at default and validation thresholds.}
\label{fig:results}
\end{figure}

\section{Discussion}
Dimensionality is associated with both performance and failure mode in our experiments. Added 3D context did not guarantee stable behavior in this imbalanced setting; higher-capacity models frequently produced degenerate predictions (all-positive/all-negative). ROC-AUC alone can mask clinical usability---a model with acceptable ranking may still be unusable due to extreme sensitivity--specificity imbalance. One plausible explanation for the transformer failures, particularly in 3D, is their well-known data-hungry nature: transformers typically require large-scale data and careful optimization to learn robust representations. Here, the combination of a limited effective sample size at the patient level and increased input dimensionality likely exacerbates this issue. In addition, the way transformers handle 3D inputs---often through tokenization or patch-based representations---may disrupt spatial continuity and reduce the effectiveness of contextual learning relative to convolutional approaches that inherently preserve local structure.

In contrast, the 2.5D representation consistently offered the best compromise in our comparison: it captures limited through-plane context while maintaining stable optimization and efficient resource usage, yielding robust discrimination and controllable operating thresholds without the instability seen in full 3D models. These findings align with recent work emphasizing the importance of representation choice and preprocessing in lung CT modeling \citep{hoq2025harnessing,hoq2026virtualeyes}. Finally, imaging is only one modality: other clinical data besides images---structured electronic health records, laboratory and physiologic indices, and multi-site biobank cohorts---offer complementary signal, and federated or multi-institutional designs may further improve robustness and generalization \citep{mu2026federated,barnes2026fluid,kolb2025renal}.

\section{Limitations}\label{sec:limit}
This is a deliberately preliminary, single-cohort study, and several factors are not fully isolated. (i) Dimensionality is our controlled variable, but architecture family, parameter count and effective capacity, optimization, and preprocessing are not perfectly matched across conditions; some observed failures may reflect these factors rather than dimensionality alone. (ii) Results come from a single patient-level split with a small positive count (20 test cases), yielding wide, overlapping confidence intervals, so model differences are not statistically separable. (iii) Absolute discrimination is modest (best ROC-AUC 0.682), consistent with single-series, fixed-protocol training rather than a fully optimized screening pipeline. (iv) LIDC-IDRI uses weak labels and serves only as supporting corroboration. We therefore present the resource--performance frontier and the failure-mode taxonomy as observations to be confirmed rather than as definitive superiority claims. A stronger version of this work would add repeated or cross-validated splits and multiple cohorts, capacity- and FLOP-matched baselines, formal statistical testing, and ablations that separate preprocessing and class-imbalance handling from input dimensionality.

\section{Conclusion}
In our comparison, the 2.5D CNN provided the most favorable balance of performance, stability, and efficiency for lung CT screening classification. Higher-dimensional models did not confer consistent benefits and often introduced instability or collapse. Subject to the limitations above, these findings motivate 2.5D as a practical default and a starting hypothesis to be validated at larger scale for class-imbalanced lung CT tasks.

We further note that the suitability of input dimensionality is likely task-dependent. While full 3D representations may be more appropriate for spatially intensive downstream tasks such as segmentation, where volumetric continuity is critical, our results suggest that for classification problems in lung cancer screening, 2D and 2.5D representations can be more effective, owing to their favorable trade-off between computational cost, model complexity, and stable learning behavior under class imbalance.

\nocite{*}
\bibliography{refs}

\clearpage
\appendix

\section*{Appendix A: Extended Results and Analysis}

\subsection*{A.1 Additional Experimental Details}
All models were trained for 20 epochs. CNNs used Adam with learning rate $10^{-4}$ and weight decay $10^{-4}$; ViTs used AdamW with learning rate $10^{-6}$ and the same weight decay. Weighted binary cross-entropy addressed class imbalance. NLST used a leakage-free patient-level split with one canonical, QC-approved CT series per patient \citep{hoq2026virtualeyes}; the QC pipeline standardizes series selection, orientation, and lung-region localization prior to input construction. The three input types are derived from the \emph{same} canonical volume and differ only in dimensionality: 2D (central axial slice through the lung region), 2.5D (three orthogonal slices stacked as channels), and 3D (a lung-centered sub-volume). CNN and ViT backbones were adapted to each input under the identical protocol above; we did not enforce exact parameter- or FLOP-matching across families, which (together with preprocessing and class-imbalance handling) is a confound we flag in Section~\ref{sec:limit}. LIDC-IDRI used XML-derived weak labels for external corroboration only.

\subsection*{A.2 Class Balance}
The task is strongly imbalanced. The held-out NLST test set ($n=297$) comprises 20 cancer-positive and 277 non-cancer patients ($\approx$6.7\% prevalence); reported sensitivity values (multiples of $1/20$) and specificity values (over 277 negatives) are consistent with these counts. Splits were constructed to preserve patient-level prevalence, and weighted BCE compensates for the imbalance during training.

\subsection*{A.3 Expanded NLST Results}
\begin{table}[htbp]
\centering
\small
\resizebox{\linewidth}{!}{%
\begin{tabular}{lcccccccc}
\toprule
Model & ROC-AUC & 95\% CI & PR-AUC & Sens@0.5 & Spec@0.5 & Sens@Val & Spec@Val & GPU MB \\
\midrule
2D CNN   & 0.581 & [0.458, 0.697] & 0.088 & 0.10 & 0.949 & 0.10 & 0.931 & 1620 \\
2.5D CNN & 0.682 & [0.546, 0.799] & 0.158 & 0.20 & 0.949 & 0.75 & 0.469 & 1646 \\
3D CNN   & 0.622 & [0.500, 0.735] & 0.107 & 0.05 & 0.975 & 0.50 & 0.671 & 1777 \\
2D ViT   & 0.598 & [0.478, 0.707] & 0.088 & 0.10 & 0.910 & 0.10 & 0.881 & 4959 \\
2.5D ViT & 0.631 & [0.515, 0.744] & 0.127 & 0.10 & 0.986 & 0.60 & 0.505 & 4959 \\
3D ViT   & 0.589 & [0.485, 0.694] & 0.081 & 0.00 & 1.000 & 0.00 & 0.964 & 352 \\
\bottomrule
\end{tabular}%
}
\caption{Expanded NLST metrics. The confidence intervals overlap across all six models, so the differences in ROC-AUC are not statistically separable on this single split.}
\label{tab:nlst_expanded}
\end{table}

\subsection*{A.4 Supporting LIDC-IDRI Results}
\begin{table}[htbp]
\centering
\small
\begin{tabular}{lccccc}
\toprule
Model & ROC-AUC & PR-AUC & Sensitivity & Specificity & GPU MB \\
\midrule
2D CNN   & 0.523 & 0.701 & 0.850 & 0.237 & 1619 \\
2.5D CNN & 0.521 & 0.797 & 0.167 & 1.000 & 555 \\
3D CNN   & 0.458 & 0.810 & 1.000 & 0.000 & 1061 \\
2D ViT   & 0.591 & 0.734 & 1.000 & 0.000 & 5217 \\
2.5D ViT & 0.645 & 0.762 & 1.000 & 0.000 & 5217 \\
\bottomrule
\end{tabular}
\caption{Supporting LIDC-IDRI results (weak labels; corroboration only).}
\label{tab:lidc}
\end{table}

\subsection*{A.5 Additional Figures}
\begin{figure}[htbp]
\centering
\includegraphics[width=0.48\linewidth]{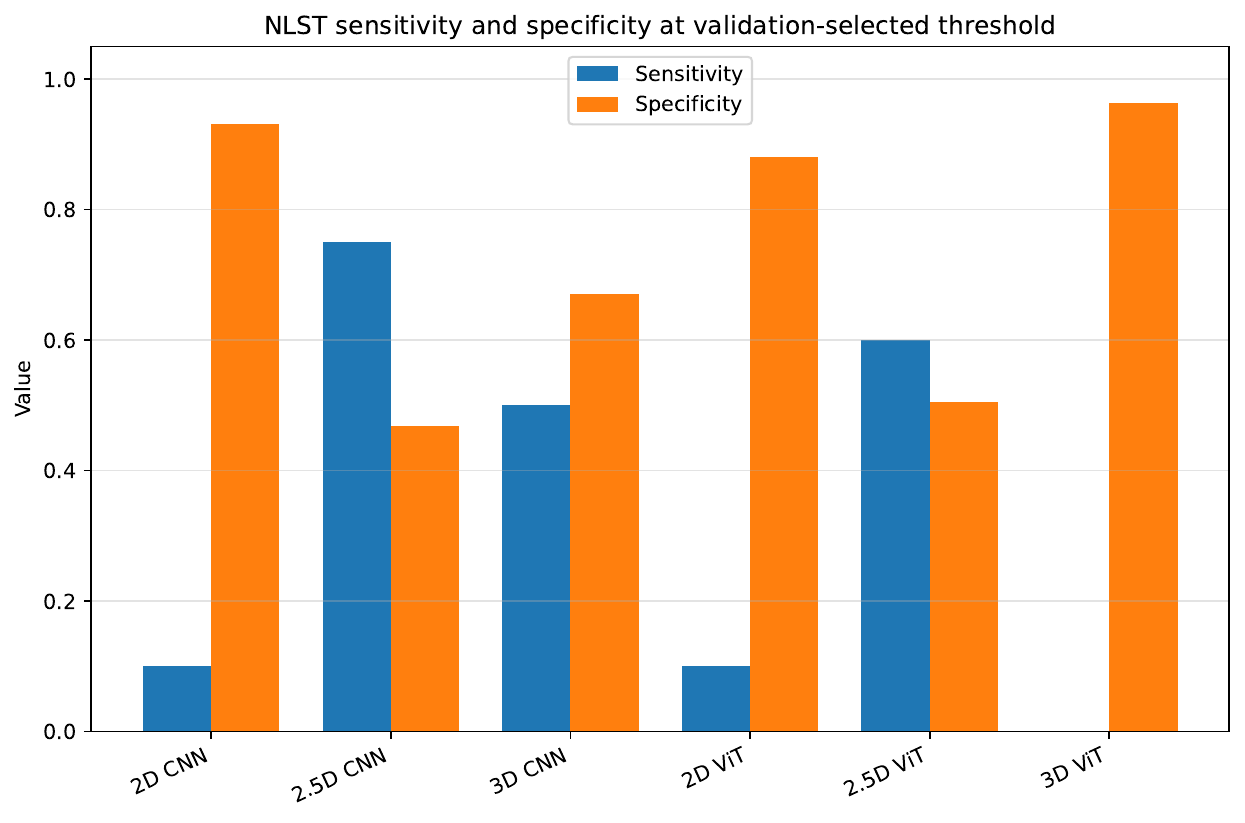}
\includegraphics[width=0.48\linewidth]{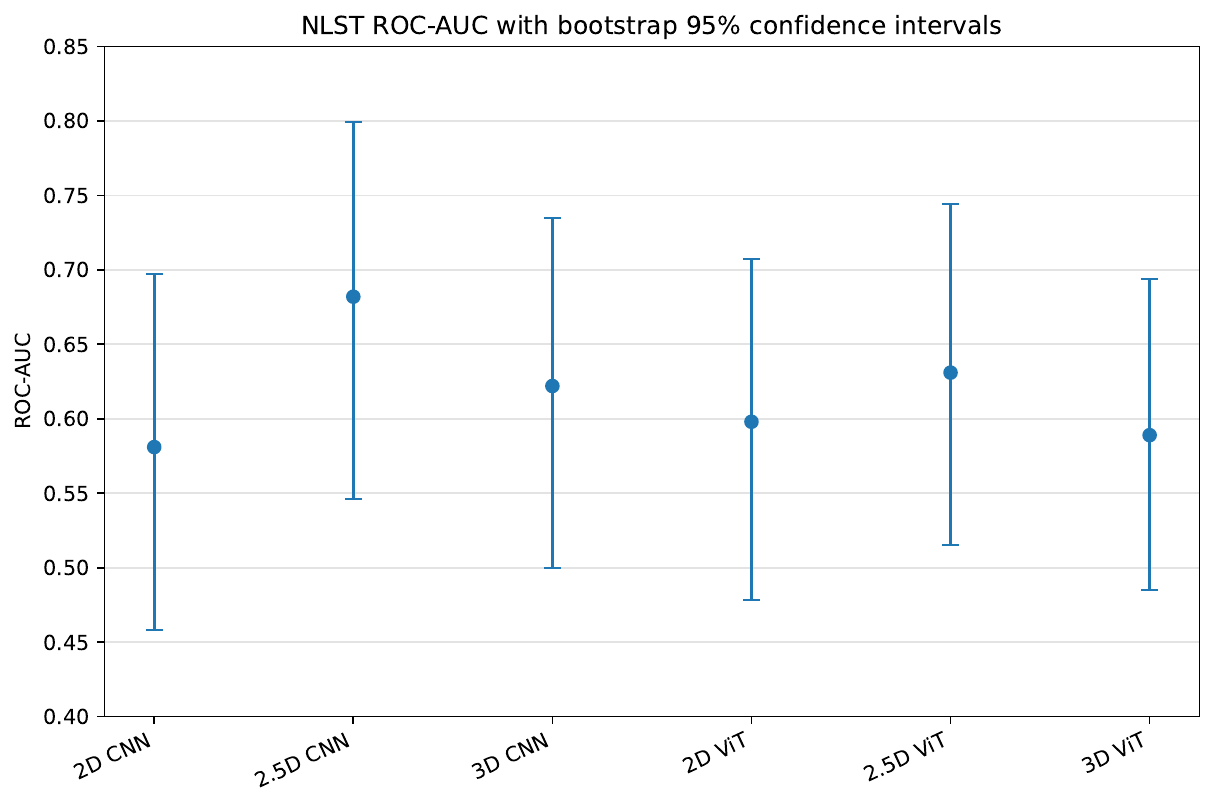}
\caption{Additional NLST evaluation plots. Left: sensitivity and specificity at the validation-selected threshold. Right: ROC-AUC with bootstrap 95\% confidence intervals.}
\label{fig:appendix_nlst}
\end{figure}

\begin{figure}[htbp]
\centering
\includegraphics[width=0.60\linewidth]{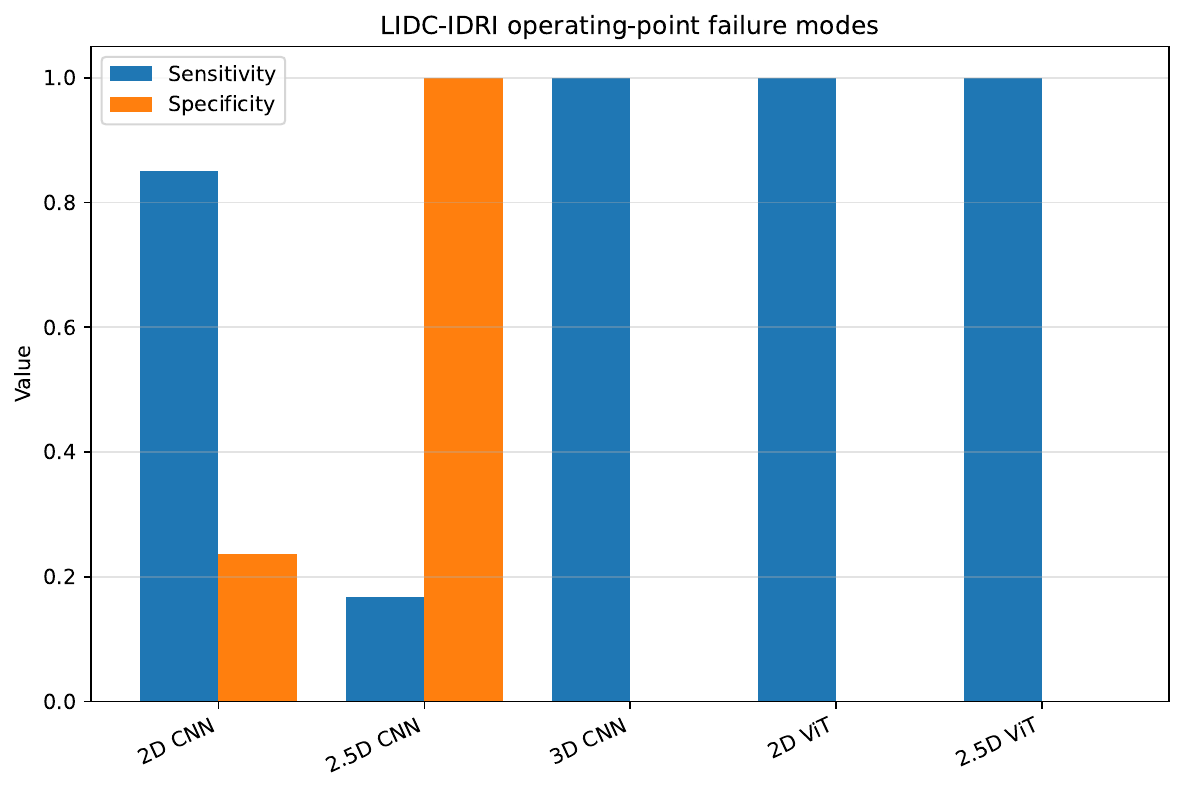}
\caption{LIDC-IDRI failure modes. Several higher-capacity models collapse to trivial all-positive behavior.}
\label{fig:appendix_lidc}
\end{figure}

\subsection*{A.6 ROC Curve Note}
Exact ROC curves for multiple models require preserved per-model prediction CSV files. The uploaded JSON summaries were sufficient for exact summary tables and confidence-interval plots, but not for reconstructing exact multi-model ROC curves after repeated uploads with the same filename. Once distinct prediction CSVs are available, exact ROC panels for 2D CNN, 2.5D CNN, and 2.5D ViT can be generated directly.

\end{document}